\documentclass[letterpaper]{article} 
\usepackage{aaai25}  
\usepackage{times}  
\usepackage{helvet}  
\usepackage{courier}  
\usepackage[hyphens]{url}  
\usepackage{graphicx} 
\usepackage{natbib}  
\usepackage{caption} 
\usepackage{algorithm}
\usepackage{algorithmic}
\usepackage{url}            
\usepackage{booktabs}       
\usepackage{amsfonts}       
\usepackage{nicefrac}       
\usepackage{microtype}      
\usepackage{xcolor}         
\usepackage{amsmath}
\usepackage{amssymb}
\usepackage{makecell}
\usepackage{multirow}
\usepackage{colortbl}
\usepackage{color}
\usepackage{arydshln}
\usepackage{subfigure}
\usepackage{enumitem}

\definecolor{mypink2}{RGB}{219, 48, 122}
\usepackage{newfloat}
\usepackage{listings}
\DeclareCaptionStyle{ruled}{labelfont=normalfont,labelsep=colon,strut=off} 
\floatstyle{ruled}
\newfloat{listing}{tb}{lst}{}
\floatname{listing}{Listing}
\title{Depth-Centric Dehazing and Depth-Estimation from \\ Real-World Hazy Driving Video}
\author{
Junkai Fan\textsuperscript{\rm 1}, Kun Wang\textsuperscript{\rm 1}, Zhiqiang Yan\textsuperscript{\rm 1}, Xiang Chen\textsuperscript{\rm 1}, Shangbing Gao\textsuperscript{\rm 2}, Jun Li\textsuperscript{\rm 1}\thanks{Corresponding authors} and Jian Yang\textsuperscript{\rm 1}\footnotemark[1]
}
\affiliations{
\textsuperscript{\rm 1}PCA Lab, Key Lab of Intelligent Perception and Systems for High-Dimensional Information of Ministry of Education, School of Computer Science and Engineering, Nanjing University of Science and Technology, China\\
\textsuperscript{\rm 2}Huaiyin Institute of Technology, China\\
\{junkai.fan, kunwang, yanzq, chenxiang, junli, csjyang\}@njust.edu.cn, gaoshangbing@hyit.edu.cn
}
\usepackage{bibentry}

\begin{document}

\maketitle

\begin{abstract}
In this paper, we study the challenging problem of simultaneously removing haze and estimating depth from real monocular hazy videos. These tasks are inherently complementary: enhanced depth estimation improves dehazing via the atmospheric scattering model (ASM), while superior dehazing contributes to more accurate depth estimation through the brightness consistency constraint (BCC). To tackle these intertwined tasks, we propose a novel depth-centric learning framework that integrates the ASM model with the BCC constraint. Our key idea is that both ASM and BCC rely on a shared depth estimation network. This network simultaneously exploits adjacent dehazed frames to enhance depth estimation via BCC and uses the refined depth cues to more effectively remove haze through ASM. Additionally, we leverage a non-aligned clear video and its estimated depth to independently regularize the dehazing and depth estimation networks. This is achieved by designing two discriminator networks: $D_\text{MFIR}$ enhances high-frequency details in dehazed videos, and $D_\text{MDR}$ reduces the occurrence of black holes in low-texture regions. Extensive experiments demonstrate that the proposed method outperforms current state-of-the-art techniques in both video dehazing and depth estimation tasks, especially in real-world hazy scenes. Project page: \url{https://fanjunkai1.github.io/projectpage/DCL/index.html }.
\end{abstract}

\section{Introduction}
Recently, video dehazing and depth estimation in real-world monocular hazy video have garnered increasing attention due to their importance in various downstream visual tasks, such as object detection~\cite{hahner2021fog}, semantic segmentation~\cite{ren2018deep}, and autonomous driving~\cite{li2023domain}. Most video dehazing methods \cite{xu2023video,fan2024driving} rely on a single-frame haze degradation model expressed through the atmospheric scattering model (ASM)~\cite{mccartney1976optics,narasimhan2002vision}:
\begin{equation}
I(x) = J(x)t(x) + A_{\infty}(1-t(x)), 
\label{eq.asm}
\end{equation}
where $I(x)$, $J(x)$ and $t(x)$ denote the hazy image, clear image and transmission map at a pixel position $x$, respectively. $A_{\infty}$ represents the infinite airlight and $t(x) = e^{-{\beta(\lambda)}d(x)}$ with $d(x)$ as the scene depth and $\beta(\lambda)$ as the scattering coefficient for wavelength $\lambda$. Clearly, $t(x)$ shows that ASM is depth-dependent, with improved depth estimation leading to better dehazing performance. Depth is estimated from real monocular hazy video using a brightness consistency constraint (BCC) between a pixel position $x$ in the current frame and its corresponding pixel position $y$ in an adjacent frame~\cite{wang2021regularizing}, that is, 
\begin{equation}
y \sim KP_{x\rightarrow y}d(x)K^{-1}x,
\label{eq.bcc}
\end{equation}
where $K$ is the camera intrinsic parameter and $P_{x\rightarrow y}$ is the relative pose for the reprojection. This suggests that clearer frames contribute to more accurate depth estimation. These two findings motivate the integration of the ASM model with the BCC constraint into a unified learning framework.

\begin{figure}[t]
	\centering
	\includegraphics[width=0.95\linewidth]{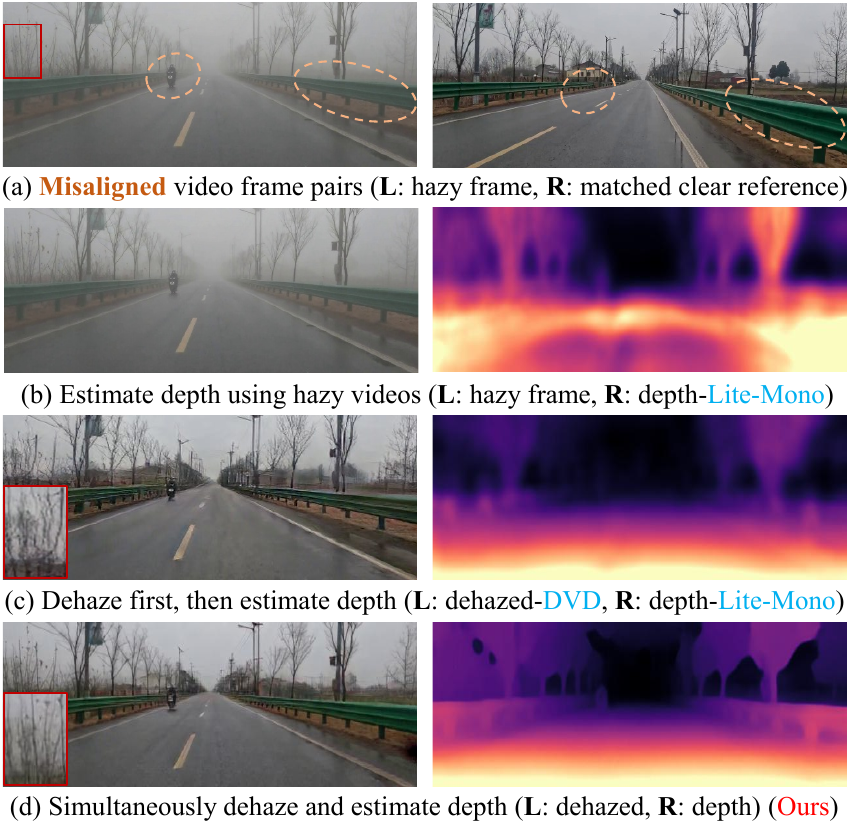}
	\caption{Visual comparisons of DVD \cite{fan2024driving}, Lite-Mono \cite{zhang2023lite} and our DCL for dehazing and depth estimation in real-world hazy scenes.} 
	\label{fig1:motivation}
\end{figure}

In practice, while both ASM and BCC yield promising dehazing results and depth estimates on synthetic hazy videos~\cite{xu2023video, gasperini2023robust}, respectively, they often fall short in real-world scenes as it is difficult to capture accurately aligned ground truth due to unpredictable weather conditions and dynamic environments \cite{fan2024driving}. For example, Fig.\ref{fig1:motivation} (b) shows the blurred depth obtained from the hazy video using Lite-Mono \cite{zhang2023lite}. To improve dehazing performance, DVD \cite{fan2024driving} introduces a non-aligned regularization (NAR) strategy that collects clear non-aligned videos to regularize the dehazing network. However, DVD still produces dehazed frames with weak textures, causing blurred depth, in Fig.~\ref{fig1:motivation} (c). 

Based on the above discussions, we propose a new Depth-Centric Learning (DCL) framework to simultaneously remove haze and estimate depth from real-world monocular hazy videos by effectively integrating the ASM model and the BCC constraint. First, starting with the hazy video frame, we design a shared depth estimation network to predict the depth $d$. Second, we define distinct deep networks to compute adjacent dehazed frames $J_x$ and $J_y$, the scattering coefficient $\beta$, and the relative pose $P_{x\rightarrow y}$, respectively. Moreover, dark channel~\cite{he2010single} is used to calculate the $A_{\infty}$ value. Third, these networks are trained using the ASM model to reconstruct the hazy frame while the BCC constraint is employed to reproject pixels from $J_x$ to $J_y$.

Inspired by the NAR strategy, we leverage a clear non-aligned video to estimate accurate depth using MonoDepth2 \cite{godard2019digging}, thereby constraining both the dehazing and depth estimation networks. Specifically, we introduce a Misaligned Frequency \& Image Regularization discriminator, $D_\text{MFIR}$, which assists the discriminator network in constraining the dehazing network to recover more high-frequency details by utilizing frequency domain information obtained through wavelet transforms~\cite{gao2021high}. Additionally, the accurate depth maps serve as references for Misaligned Depth Regularization discriminator $D_\text{MDR}$, further mitigating issues such as black holes in depth maps caused by weak texture regions. Our contributions are summarized as follows:
\begin{itemize}[leftmargin=*] 
\item To the best of our knowledge, we are the first to propose a Depth-centric Learning (DCL) framework that effectively integrates the atmospheric scattering model and brightness consistency constraint, enhancing both video dehazing and depth estimation simultaneously. 
\item We introduce two discriminator networks, $D_\text{MFIR}$ and $D_\text{MDR}$, to address the loss of high-frequency details in dehazed images and the black holes in depth maps with weak textures, respectively.
\item We evaluate the proposed method separately using video dehazing datasets (e.g., GoProHazy, DrivingHazy and InternetHazy) \cite{fan2024driving} and depth estimation datasets (e.g., DENSE-Fog) \cite{bijelic2020seeing} in real hazy scenes. The experimental results demonstrate that our method exceeds previous state-of-the-art competitors.
\end{itemize}

\section{Related work}

\textbf{Image/video dehazing.} Early methods for image dehazing primarily focused on integrating atmospheric scattering models (ASM) with various priors~\cite{he2010single,fattal2014dehazing}, while recent advances have leveraged deep learning with large hazy/clear image datasets~\cite{li2018benchmarking,wenxuan2025SGDN}. These approaches use neural networks to either learn physical model parameters~\cite{deng2019deep,li2021you,liu2022towards} or directly map hazy to clear images/videos~\cite{qu2019enhanced,qin2020ffa,ye2022perceiving}. However, they rely on aligned synthetic data, resulting in domain shifts in real-world scenarios. To address this, domain adaptation~\cite{shao2020domain,chen2021psd,wu2023ridcp} and unpaired dehazing model~\cite{zhao2021refinednet,yang2022self,wang2024odcr} are used for real dehazing scenes. Despite these efforts, image dehazing models still encounter brightness inconsistencies between adjacent frames when applied to videos, leading to noticeable flickering.

Video dehazing techniques leverage temporal information from adjacent frames to enhance restoration quality. Early methods focused on post-processing to ensure temporal consistency by refining transmission maps~\cite{ren2018deep} and suppressing artifacts~\cite{chen2016robust}. Some approaches also addressed multiple tasks, such as depth estimation~\cite{li2015simultaneous} and detection~\cite{li2018end}, within hazy videos. Recently, \cite{zhang2021learning} introduced the REVIDE dataset and a confidence-guided deformable network, while~\cite{liu2022phase} proposed a phase-based memory network. Similarly,~\cite{xu2023video} developed a memory-based physical prior guidance module for incorporating prior features into long-term memory. Although some image restoration methods~\cite{yang2023video} excel on REVIDE in adverse weather, they are mainly trained on indoor smoke scenes, limiting their effectiveness in real outdoor hazy conditions.

In contrast to previous dehazing and depth estimation works~\cite{yang2022self,chen2023dehrformer}, our DCL is trained on real hazy video instead of synthetic hazy images. Furthermore, it simultaneously optimizes both video dehazing and depth estimation by integrating the ASM model with the brightness consistency constraint (BCC).

\begin{figure*}[t]
	\centering
	\includegraphics[width=0.85\linewidth]{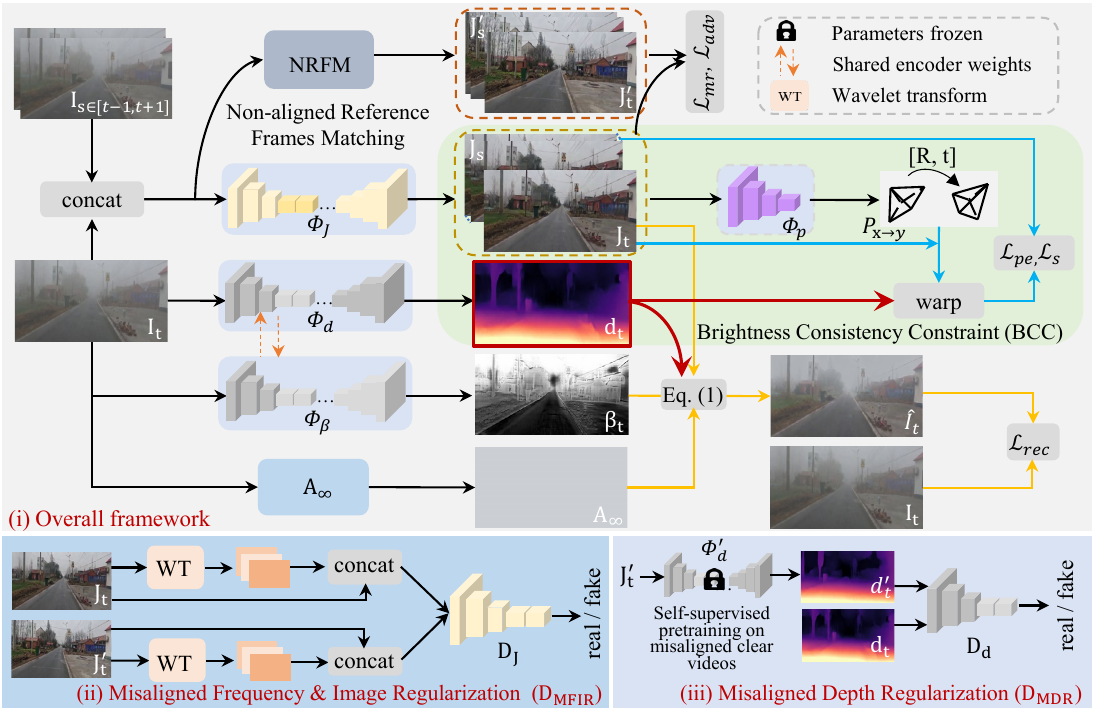}
	\caption{The pipeline of our Depth-Centric Learning (DCL) framework. It effectively integrates the atmospheric scattering model with the brightness consistency constraint through shared depth prediction. $D_\text{MFIR}$ enhances high-frequency detail recovery in dehazed frames, while $D_\text{MDR}$ reduces black holes in depth maps caused by weakly textured regions.}
	\label{fig2: ovaerall framework}
\end{figure*}

\noindent\textbf{Self-supervised Monocular Depth Estimation (SMDE).} 
Laser radar perception is limited in extreme weather, leading to a growing interest in self-supervised methods. Building on the groundbreaking work \cite{zhou2017unsupervised}, which showed that geometric constraints between consecutive frames can achieve strong performance, researchers have explored various cues for self-supervised training using video sequences \cite{godard2019digging, watson2021temporal, liu2024mono} or stereo image pairs \cite{godard2017unsupervised}. Recently, advanced networks \cite{zhang2023lite, zhao2022monovit} have been used for self-supervised depth estimation (SMDE) in challenging conditions like rain, snow, fog, and low light. However, degraded images, especially in low-texture areas, significantly hinder accurate depth estimation. To address this,   some methods focus on image enhancement \cite{wang2021regularizing, zheng2023steps} and domain adaptation \cite{gasperini2023robust, saunders2023self}. While these methods improve depth estimation, they typically rely on operator-based enhancements rather than learnable approaches. Additionally, domain adaptation, often based on synthetic data, may not generalize well to real-world scenes.

Compared to the above SMDE methods, our approach leverages a learnable framework based on a physical imaging model, trained directly on real-world data. This approach effectively mitigates the domain gap, offering improved performance in real-world scenarios. 

\section{Methodology}
In this section, we propose a novel Depth-centric Learning (DCL) framework (see Fig.~\ref{fig2: ovaerall framework}) to simultaneously remove haze and estimate depth from real-world monocular hazy videos. First, we introduce a unified ASM-BCC model that effectively integrates the ASM model and the BCC constraint. Next, we present two misaligned regularization discriminator networks, $D_\text{MFIR}$ and $D_\text{MDR}$, for enhancing constraints on high-frequency details and weak texture regions. Finally, we outline the overall training loss.

\subsection{A Unified ASM-BCC Model}\label{Sec 3.1}
For a given hazy video clip with a current frame $I_{t}$ and its adjacent frames, $I_{s\in[t-n:t+n],s\neq t}\in\mathbb{R}^{3\times h\times w}$, where $h$ is the height, $w$ is the width and we set $n=1$, we define a unified ASM-BCC model by combining Eqs. \eqref{eq.asm} and \eqref{eq.bcc}:
\begin{align}
\left\{\begin{array}{ll}
I_t(x) = J_t(x)e^{-{\beta}\textbf{d(x)}} + A_{\infty}\left(1-e^{-{\beta}\textbf{d(x)}}\right) \\
J_{t}(x)=\mathcal{S}(J_s, y), \ y \sim KP_{x\rightarrow y}\textbf{d(x)}K^{-1}x \\
\end{array}
\right.
\label{eq-ASM-BCC} 
\end{align}
where $J_{s}$ is a dehazed frame computed by the dehazing network from the hazy frame $I_s$. The function $\mathcal{S}$ denotes the differentiable bilinear sampling operation~\cite{jaderberg2015spatial}, $x$ and $y$ represent the pixel positions of the frames $I_{t}$ and $I_{s\in[t-n:t+n],s\neq t}$, respectively. $K$ denotes the camera's intrinsic parameters. Then, we define various networks to predict the variables in the Eq.~\eqref{eq-ASM-BCC}. Starting from the inputs $I_t$ and $I_{s}$, we design a dehazing network $\varPhi_J(I_{[t-n:t+n]})$ to recover $[J_t,J_s]$, a shared depth estimation network $\varPhi_d(I_t)$ to predict $d\in\mathbb{R}^{1\times h\times w}$, and a scattering coefficient estimation network $\varPhi_\beta(I_t)$ to learn $\beta\in\mathbb{R}^{1\times h\times w}$, a pose estimation network $\varPhi_p(J_t, J_s)$ to predict the relative pose $P_{x\rightarrow y}\in\mathbb{R}^{4\times4}$, respectively. Additionally, the infinite airlight $A_{\infty}$ is calculated by taking the mean of the brightest 1\% pixels from the dark channel~\cite{he2010single}. 
From Eq. \eqref{eq-ASM-BCC}, it is clear that the ASM model and the BCC constraint are seamlessly and logically integrated into a unified ASM-BCC model. This integration enables us to perform two critical tasks simultaneously: video dehazing and depth estimation. 

\textbf{Remark:} In real hazy scenes, scattering does not always conform to an ideal model, as $\beta$ depends not only on wavelength but also on the size and distribution of scattering particles (e.g., patchy haze)~\cite{mccartney1976optics,zhou2021learning}. Therefore, we assume $\beta$ to be a non-uniform variable.

Next, we introduce the SAM and BCC loss functions employed to train our ASM-BCC model.

\textbf{ASM Loss}. According to the upper equation in Eq. \eqref{eq-ASM-BCC}, $I_t$ can be reconstructed using $[J_t,J_s]=\varPhi_J(I_{[t-n:t+n]})$, $d_t=\varPhi_d(I_t)$, $\beta=\varPhi_\beta(I_t)$, and $A_{\infty}$. Following previous works \cite{fan2023non,fan2024driving}, employ a reconstruction loss $\mathcal{L}_{rec}$ to supervise the learning of these three variables from numerical, structural, and perceptual perspectives. $\mathcal{L}_{rec}$ is formulated as
\begin{align}
	\mathcal{L}_{rec}(I_t, \widehat I_{t}) = ||I_t-\widehat I_{t}||_1 + \mathbb{S}(I_t, \widehat I_{t}) + \mathbb{P}(I_t, \widehat I_{t}),
	\label{eq:rec_loss}
\end{align}
where $\widehat I_{t}=J_te^{-\beta_td_t} + A_{\infty}\left(1-e^{-\beta_td_t}\right)$, $\mathbb{S}$~\cite{wang2004image} and $\mathbb{P}$~\cite{johnson2016perceptual} are the measures of structural and perceptual similarity, respectively.

To mitigate the difficulty of obtaining strictly aligned ground truth, we employ Non-aligned Reference Frames Matching (NRFM)~\cite{fan2024driving} to identify a non-aligned clear video frame $J'_{t}$ from the same scene. This frame is used to supervise the current dehazed frame $J_t$ during the training of the dehazing network $[J_t,J_s]=\varPhi_J(I_{[t-n:t+n]})$. The corresponding regularization is defined as:
\begin{align}
	\mathcal{L}_{mr}(J_t, J'_t) = \sum\nolimits_{l=1}^{5}\mathcal{D}(\varOmega^l(J_t),\varOmega^l(J'_t)),
	\label{eq:mr_loss}
\end{align}
where $\mathcal{D}(\cdot, \cdot$) represents the cosine distance between $J_t$ and $J'_t$ in the feature space. $\varOmega^l(J_t)$ and $\varOmega^l(J'_t)$ denote the feature maps extracted from the $l$-th layer of VGG-16 network with inputs $J_t$ and $J'_t$, respectively.

\textbf{BCC Loss}. According to the lower equation in Eq. \eqref{eq-ASM-BCC}, we can establish a brightness consistency constraint between a pixel point $x$ in $I_t$ and its corresponding pixel point $y$ in $I_s$.This allows us to reconstruct the target frame $J_t$ from $J_s$ using the parameter $K$, the networks $d_t=\varPhi_d(I_t)$, $[J_t,J_s]=\varPhi_J(I_{[t-n:t+n]})$, and $P_{x\rightarrow y}=\varPhi_p(J_t, J_s)$. Following the approach in~\cite{godard2017unsupervised},  we combine the $\ell_1$ distance and structural similarity together as the photometric error for the BCC loss,
\begin{align}
\resizebox{.9\hsize}{!}{$
\mathcal{L}_{pe}(\widehat J_t, J_t) = \frac{\alpha}{2}(1-\mathbb{S}(\widehat J_t, J_t)) + (1-\alpha)||\widehat J_t - J_t||_1,$} 
	\label{eq:pe_loss}\\
\text{with} \ m_a  = [\mathcal{L}_{pe}(J_t, \widehat J_t) < \mathcal{L}_{pe}(J_t, J_s)], \ \ \ \ \ \ \ \ \ 
\label{eq:mask}
\end{align}
where $\widehat J_t = \mathcal{S}(J_s, y)$, $y \sim KP_{x\rightarrow y}d_tK^{-1}x$, $\mathbb{S}$ represents the SSIM loss, $\mathcal{S}$ denotes the differentiable bilinear sampling operation~\cite{jaderberg2015spatial}. Throughout all experiments, we set $\alpha = 0.85$ in all experiment. The term $m_a$ refers to a mask generated using the auto-mask strategy~\cite{godard2019digging}. Additionally, to address depth ambiguity, we apply an edge-aware smoothness loss~\cite{godard2017unsupervised} to enforce depth smoothness,
\begin{align}
	\mathcal{L}_{s}(d^*_t, J_t) = |\partial_xd^*_t|e^{-|\partial_xJ_t|} + |\partial_yd^*_t|e^{-|\partial_yJ_t|},
	\label{eq:s_loss}
\end{align}
where $d^*_t$ represents the mean-normalized inverse depth. The operators $\partial_x$ and $\partial_y$ denote the image gradients along the horizontal and vertical axes, respectively.

\subsection{Misaligned Regularization}\label{Sec 3.2} 

\textbf{\emph{Misaligned Frequency \& Image Regularization (MFIR).}} To ensure that the dehazing network $\varPhi_J$ produces results with rich details, we regularize its output using misaligned clear reference frames via the proposed MFIR discriminator network $D_\text{MFIR}$. For this, we adopt a classic wavelet transformation technique, specifically the Haar wavelet, which involves two operations: wavelet pooling and unpooling. Initially, wavelet pooling is applied to extract high-frequency features $\mathbb{F}_\text{LH}$, $\mathbb{F}_\text{HL}$ and $\mathbb{F}_\text{HH}$, which are then concatenated with the image and passed into $D_\text{MFIR}$. This strategy promotes the generation of dehazed images with enhanced high-frequency components, thus improving their visual realism. The adversarial loss for $D_\text{MFIR}$ is defined as follows:
\begin{equation}
\begin{split}
\mathcal{L}^{I}_{D} =& \mathbb{E}[\log(D_J(cat(\mathbb{F}'_{[\text{LH,HL,HH}]}, J'_t))-1)^2], \\
&+ \mathbb{E}[\log(D_J(cat(\mathbb{F}_{\text{[LH,HL,HH}]}, J_t))^2] \\  
\mathcal{L}^{I}_{G} =& \mathbb{E}[\log(D_J(cat(\mathbb{F}_{[\text{LH,HL,HH}]}, J_t))-1)^2], 
\label{eq:MFIR}
\end{split}
\end{equation}
where $cat(\cdot, \cdot)$ denotes the concatenation operation along channel dimension, $D_J$ represents a discriminator network. $J'_t$ and $J_t$ refer to misaligned reference frames and the corresponding dehazing results, respectively. 

\textbf{\emph{Misaligned Depth Regularization (MDR).}}
We further extend the above misaligned regularization to address weak texture issues in self-supervised depth estimation within the depth estimation network. To obtain high-quality reference depth maps, we train a depth estimation network $\varPhi'_d$ to produce $d'_t$ in a self-supervised manner using MonoDepth2 \cite{godard2019digging} from clear misaligned video frames. In comparison to the unpaired regularization approach in~\cite{wang2021regularizing}, misaligned regularization enforces a more stringent constraint. The optimization objective for 
$D_\text{MDR}$ can be formulated as follows:
\begin{equation}
\begin{split}
\mathcal{L}^{d}_{D} =& \mathbb{E}[\log(D_d(\mu(d'_t))-1)^2] \\
&+ \mathbb{E}[\log(D_d(\mu(d_t))^2], \\  
\mathcal{L}^{d}_{G} =& \mathbb{E}[\log(D_d(\mu(d_t))-1)^2],
\label{eq:MDR}
\end{split}
\end{equation}
where the depth normalization, $\mu(d) = d / \bar{d}$, eliminates scale ambiguity by dividing the depth $d$ by its mean $\bar{d}$. This step is essential because both $d_t$ and $d'_t$ exhibit scale ambiguity, making direct scale standardization unreasonable.

\begin{table*}[!t]
	\centering
	\setlength\tabcolsep{2.3pt}
	\renewcommand\arraystretch{1.2}
	\scalebox{0.90}{
		\begin{tabular}{c|c|c||c|c||c|c||c|c||c|c|c||c}
			\hline
			\multicolumn{1}{c|}{\multirow{2}[1]{*}{\makecell{Data\\Settings}}}
			& \multicolumn{1}{c|}{\multirow{2}[1]{*}{Methods}}  
			& \multicolumn{1}{c||}{\multirow{2}[1]{*}{\makecell{Data\\Type}}} 
			& \multicolumn{2}{c||}{GoProHazy}  
			& \multicolumn{2}{c||}{DrivingHazy}  
			& \multicolumn{2}{c||}{InternetHazy} 
			&\multirow{2}[1]{*}{\makecell{Params\\(M)}} 
			&\multirow{2}[1]{*}{\makecell{FLOPs\\(G)}} 
			&\multirow{2}[1]{*}{\makecell{Inf. time\\(S)}} 
			&\multirow{2}[1]{*}{Ref.}\\
			\cline{4-9}     &  &  
			& \multicolumn{1}{c|}{FADE $\downarrow$} & \multicolumn{1}{c||}{NIQE $\downarrow$} 
			& \multicolumn{1}{c|}{FADE $\downarrow$} & \multicolumn{1}{c||}{NIQE $\downarrow$} 
			& \multicolumn{1}{c|}{FADE $\downarrow$} & \multicolumn{1}{c||}{NIQE $\downarrow$}  &  &  & \\
			\hline
			\multirow{4}[2]{*}{Unpaired} 			
			& DCP 		   & Image  & 1.0415  & 7.4165  & 1.1260  & 7.4455   & 0.9229  & 7.4899   &-        & -      & 1.39	& CVPR'09  \\
			& RefineNet    & Image  & 1.1454  & 6.1837  & 1.0223  & 6.5959   & 0.8535  & 6.7142   & 11.38   & 75.41  & 0.105	& TIP'21  \\
			& CDD-GAN 	   & Image  & 0.7797  & 6.0691  & 1.0072  & 6.1968   & 0.8166  & 6.1969   & 29.27   & 56.89  & 0.082	& ECCV'22  \\
			& D$^{4}$ 	   & Image  & 1.5618  & 6.9302  & 0.9556  & 7.0448   & 0.6913  & 7.0754   & {\bf 10.70}   & {\bf 2.25}   & 0.078	& CVPR'22  \\
			\hdashline
			\multirow{2}[9]{*}{Paired} 
			& PSD 		   & Image  & 0.9081  & 6.7996  & 0.9479  & 6.3381   & 0.8100  & 6.1401   & 33.11   & 182.5  & 0.084	& CVPR'21  \\
			& RIDCP 	   & Image  & 0.7250  & 5.2559  & 0.9187  & 5.3063   & 0.6564  & 5.4299   & 28.72   & 182.69 & 0.720	& CVPR'23  \\
			& PM-Net 	   & Video  & 0.7559  & 4.6274  & 1.0509  & 4.8447   & 0.7696  & 5.0182   & 151.20  & 5.22   & 0.277	& ACMM'22  \\
			& MAP-Net 	   & Video  & 0.7805  & 4.8189  & 1.0992  & 4.7564   & 1.0595  & 5.5213   & 28.80   & 8.21   & 0.668	& CVPR'23  \\
			\hdashline
			\multirow{2}[5]{*}{Non-aligned} 
			& NSDNet 	   & Image  & 0.7197  & 6.1026  & 0.8670  & 6.3558   & 0.6595  & 4.3144   & 11.38   & 56.86  & {\bf 0.075} 	& arXiv'23  \\
			& DVD          & Video  & 0.7061  & 4.4473  & 0.7739  & 4.4820   & 0.6235  & 4.5758   & 15.37   & 73.12  & 0.488	& CVPR'24  \\
			& \textbf{DCL (Ours)} & Video  &{\bf 0.6914} &{\bf 3.4412} &{\bf 0.7380} &{\bf 3.5329} &{\bf 0.6203} &{\bf 3.5545} &11.38 &56.86 &{\bf 0.075} &- \\
			\hline
	\end{tabular}}
	\caption{Quantitative dehazing results on three real-world hazy video datasets. The symbol $\downarrow$ denotes that lower values are better.  \emph{Note that all quantitative evaluations were performed at an output resolution of 640$\times$192.} } 
	\label{tab1:dehaze}%
\end{table*}%

\begin{figure*}[!t]
	\centering
	\subfigure{
		\begin{minipage}[b]{0.97\textwidth}	
			\includegraphics[width=1\textwidth]{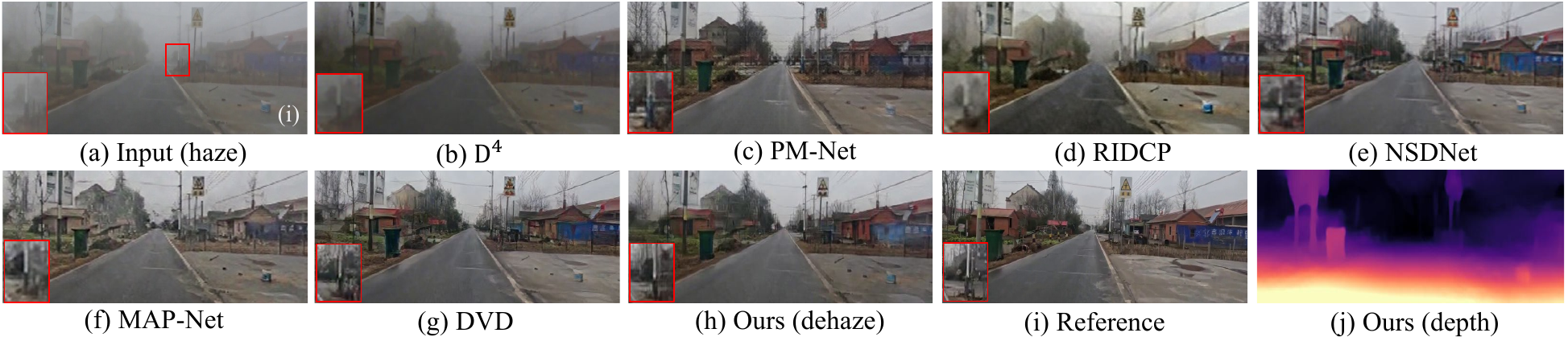}	
	\end{minipage}}
	\subfigure{
		\begin{minipage}[b]{0.97\textwidth}
			\includegraphics[width=1\textwidth]{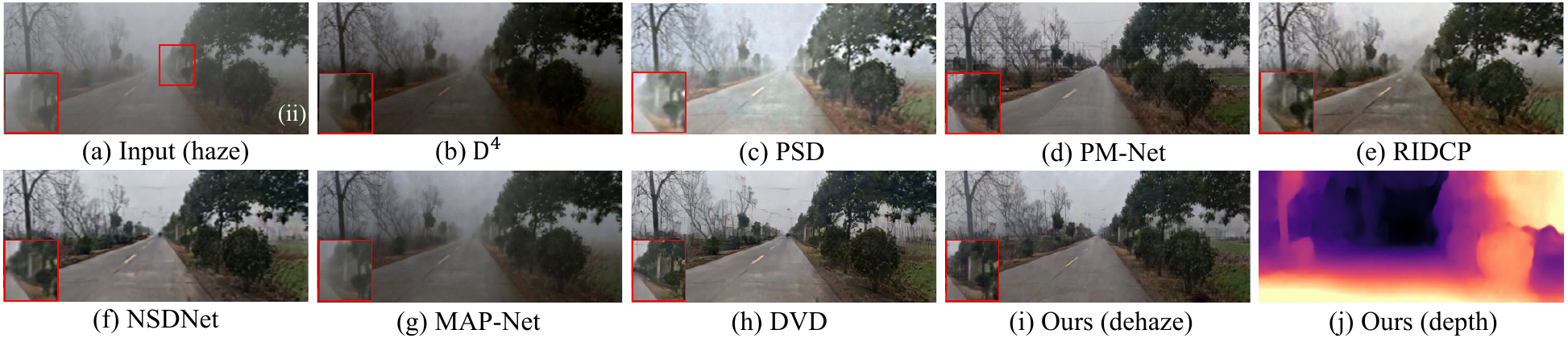}
	\end{minipage}}
	\subfigure{
		\begin{minipage}[b]{0.97\textwidth}
			\includegraphics[width=1\textwidth]{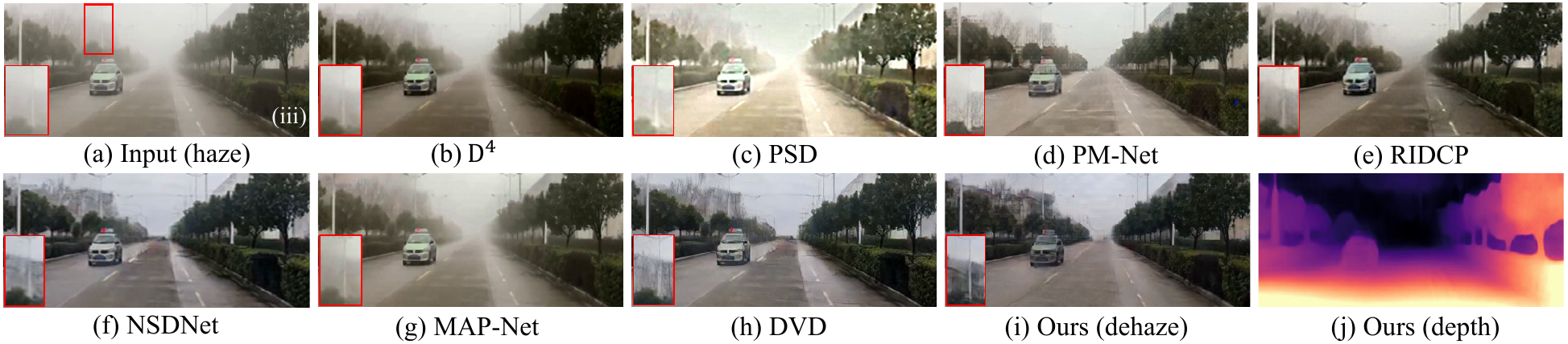}
	\end{minipage}}
	\caption{Comparisons of video dehazing performance across (i) GoProHazy, (ii) DrivingHazy, and (iii) InternetHazy. Our method effectively removes haze and accurately estimates depth. The red box highlights a zoomed-in region for clearer comparison.}
	\label{fig3:dehaze}
\end{figure*}

\subsection{Overall Training Loss}\label{train_loss}

The final loss is composed of several terms: the reconstruction loss in Eq.~\eqref{eq:rec_loss}, the misaligned reference loss in Eq.~\eqref{eq:mr_loss}, the photometric loss in Eq.~\eqref{eq:pe_loss}, the edge-aware smoothness loss in Eq.~\eqref{eq:s_loss}, the MFIR loss in Eq.~\eqref{eq:MFIR} and the MDR loss in Eq.~\eqref{eq:MDR}, as defined below:
\begin{align}
	\begin{split}
		Loss &= \eta\mathcal{L}_{rec} + \gamma\mathcal{L}_{mr} + m_a\mathcal{L}_{pe} + \xi\mathcal{L}_{s}  \\
		&+ \omega_1(\mathcal{L}^I_{D}+ \mathcal{L}^I_{G}) + \omega_2(\mathcal{L}^d_{D} + \mathcal{L}^d_{G}),
		\label{eq:all_loss}
	\end{split}
\end{align}
where $\eta$, $\gamma$, $\xi$, $\omega_1$ and $\omega_2$ are weight parameters and the mask $m_a$ is defined in Eq.~\eqref{eq:mask}.

\section{Experiment Results}\label{experiment results}

In this section, we evaluate the effectiveness of our proposed method by conducting experiments on four real-world hazy video datasets: GoProHazy, DrivingHazy, InternetHazy, and DENSE-Fog (which includes sparse depth ground truth). We compare our method against state-of-the-art image/video dehazing and depth estimation techniques. Additionally, we perform ablation studies to highlight the impact of our core modules and loss functions. \emph{Note that more implementation details, visual results, ablation studies, discussions and a video demo are provided in Supplemental Material.}

\begin{table*}[t]
	\centering
	\setlength\tabcolsep{1.8pt}
	\renewcommand\arraystretch{1.2}
	\scalebox{0.88}{
	\begin{tabular}{l||c|c|c|c|c||c|c|c|c|c||c|c|c||c}
        \hline
        \multicolumn{1}{l||}{\multirow{2}[1]{*}{Method}}
        & \multicolumn{5}{c||}{DENSE-Fog (light)}  
        & \multicolumn{5}{c||}{DENSE-Fog (dense)} 
        &\multirow{2}[1]{*}{\makecell{Params\\(M)}} 
        &\multirow{2}[1]{*}{\makecell{FLOPs\\(G)}} 
        &\multirow{2}[1]{*}{\makecell{Inf. time\\(S)}} 
        &\multirow{2}[1]{*}{\makecell{Ref.}} \\
        \cline{2-11} 
        & \multicolumn{1}{c|}{abs Rel$\downarrow$}     
        & \multicolumn{1}{c|}{RMSE log$\downarrow$} 
        & \multicolumn{1}{c|}{$\delta_1$$\uparrow$} 
        & \multicolumn{1}{c|}{$\delta_2$$\uparrow$} 
        & \multicolumn{1}{c||}{$\delta_3$$\uparrow$} 
        & \multicolumn{1}{c|}{abs Rel$\downarrow$}  
        & \multicolumn{1}{c|}{RMSE log$\downarrow$}  
        & \multicolumn{1}{c|}{$\delta_1$$\uparrow$}
        & \multicolumn{1}{c|}{$\delta_2$$\uparrow$} 
        & \multicolumn{1}{c||}{$\delta_3$$\uparrow$} & & &\\
        \hline
        MonoDepth2 	 &0.418  &0.475  &0.499  &0.735  &0.847  &1.045  &0.632  &0.530  &0.771  &0.864  &14.3 &8.0  &{\bf0.009}   &ICCV'19  \\
        MonoViT      &0.393  &0.454  &0.464  &0.728  &0.858  &0.992  &0.611  &0.512  &0.779  &0.876  &78.0 &15.0 &0.045   &3DV'22   \\
        Lite-Mono    &0.417  &0.473  &0.402  &0.687  &0.853  &0.954  &0.604  &0.469  &0.756  &0.886  &{\bf 3.1}  &{\bf5.1}&0.013   &CVPR'23  \\
        RobustDepth  &0.316  &0.370  &0.611  &0.828  &0.913  &{\bf0.605}  &{\bf0.515}  &0.563  &0.798  &0.881  &14.3       &8.0     &{\bf0.009}  &ICCV'23  \\
        Mono-ViFI    &0.369  &0.459  &0.408  &0.704  &0.864  &0.609  &0.528  &0.489  &0.771 &0.883    &14.3    &8.0   &{\bf0.009}        &ECCV'24  \\
        \textbf{DCL (Ours)} &{\bf0.311} &{\bf0.364} &{\bf0.623}  &{\bf0.839}  &{\bf0.920} &1.182 &0.596 &{\bf0.612} &{\bf0.829} &{\bf0.900} &14.3 &8.0  &\bf{0.009} &- \\
        \hline
	\end{tabular}}
	\caption{Quantitative depth estimation results on DENSE-Fog dataset. All methods were trained using the GoProHazy dataset. The symbols $\downarrow$ and $\uparrow$ denote that lower or higher values are better, respectively.} 
	\label{tab2:depth}
\end{table*}

\begin{figure*}[t]
	\centering
	\includegraphics[width=0.97\linewidth]{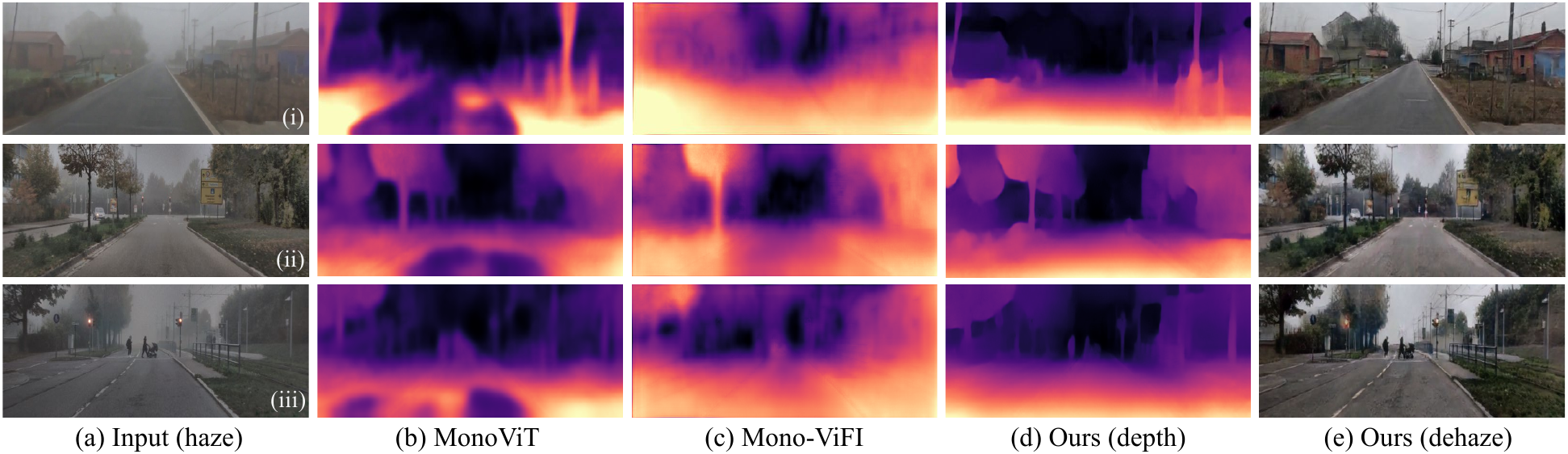}
    \vskip -0.05in
	\caption{Visual results on GoProHazy (i) and DENSE-Fog (ii-dense, iii-light). They demonstrate that our method achieves strong dehazing generalization and provides more accurate depth estimation in real hazy scenes.} 
	\label{fig4:depth}
    \vskip -0.1in
\end{figure*}

\subsection{Datasets and Evaluation Metrics}

\textbf{Three real-world video dehazing dataset \cite{fan2024driving}.}
\emph{GoProHazy}, consists of videos recorded with a GoPro 11 camera under hazy and clear conditions, comprising 22 training videos (3791 frames) and 5 testing videos (465 frames). Each hazy video is paired with a clear non-aligned reference video, with the hazy-clear pairs captured by driving an electric vehicle along the same route, starting and ending at the same points. In contrast, \emph{DrivingHazy} was collected using the same GoPro camera while driving a car at relatively high speeds in real hazy conditions. This dataset contains 20 testing videos (1807 frames), providing unique insights into hazy conditions encountered during high-speed driving. \emph{InternetHazy} contains 328 frames sourced from the internet, showcasing hazy data distributions distinct from those of \emph{GoProHazy} and \emph{DrivingHazy}. All videos in these datasets are initially recorded at a resolution of 1920$\times$1080. After applying distortion correction and cropping based on the intrinsic parameters $K$ of the GoPro 11 camera (calibrated by us), the resolutions of \emph{GoProHazy} and \emph{DrivingHazy} are 1600$\times$512.

\noindent\textbf{One real-world depth estimation dataset.} We select hazy data labeled as dense-fog and light-fog from the DENSE dataset~\cite{bijelic2020seeing} for evaluation, excluding nighttime scenes. Specifically, we used 572 dense-fog images and 633 light-fog images to assess all depth estimation models. Sparse radar points were used as ground truth for depth evaluation, with errors considered only at the radar point locations. This dataset has a resolution of 1920$\times$1024. For consistency with GoProHazy, we cropped the RGB images and depth ground truth to 1516$\times$486, maintaining a similar aspect ratio.

\noindent\textbf{Evaluation metrics.} In this work, we use FADE~\cite{choi2015referenceless} and NIQE~\cite{mittal2012making} to assess the dehazing performance. For depth evaluation, we compute the seven standard metrics-Abs Rel, Sq Rel, RMSE, RMSE log, $\delta_1<1.25$, $\delta_2<1.25^2$, $\delta_3<1.25^3$)-as proposed in~\cite{eigen2014depth} and commonly used in depth estimation tasks.

\subsection{Implementation details}\label{implementation details}

In the training process, we use the ADAM optimizer~\cite{kingma2014adam} with default parameters ($\beta_1 = 0.9$, $\beta_2 = 0.99$) and a MultiStepLR scheduler. The initial learning rate is set to $1e^{-4}$ and decays by a factor of 0.1 every 15 epochs. The batch size is 2, and the input frame size is 640$\times$192. Our model is trained for 50 epochs using PyTorch on a single NVIDIA RTX 4090 GPU, with training taking approximately 15 hours on the GoProHazy dataset. The final loss parameters are set as follows: $\eta=1e^{-1}$, $\gamma=2e^{-1}$, $\xi=1e^{-3}$, $\omega_1=4e^{-3}$ and $\omega_2=1e^{-3}$. The encoder for depth estimation, $\varPhi_d$, the scattering coefficient network, $\varPhi_\beta$, and the pose network, $\varPhi_p$, all use a ResNet-18 architecture, with $\varPhi_d$ and $\varPhi_\beta$ sharing encoder weights. The depth encoder $\varPhi'_d$ is trained using MonoDepth~\cite{godard2019digging}.

\begin{figure*}[!t]
	\centering
	\includegraphics[width=0.97\linewidth]{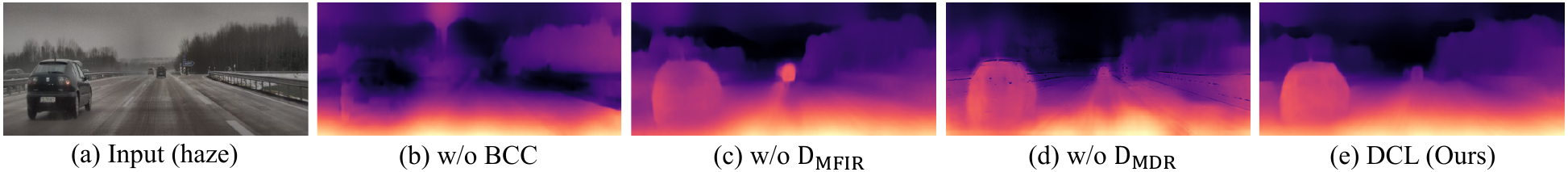}
	\caption{Ablation visualization of BCC, $D_{\text{MFIR}}$ and $D_{\text{MDR}}$ on DENSE-Fog (light).} 
    \vskip -0.05in
	\label{fig5:ablation_module}
    \vskip -0.1in
\end{figure*}

\subsection{Compare with SOTA Methods}\label{compare SOTA}

\textbf{Image/video dehazing.} 
We first evaluate the proposed DCL model on three dehazing benchmarks: GoProHazy, DrivingHazy, and InternetHazy. Notably, during testing, our DCL model relies solely on the dehazing subnetwork.  The results under unpaired, paired, and non-aligned settings are summarized in Table~\ref{tab1:dehaze}. Overall, DCL ranks \textbf{1st} in both FADE and NIQE across all methods. For instance, DCL's NIQE score outperforms the second-best non-aligned method, DVD~\cite{fan2024driving}, by \textbf{22.04\%}, and it surpasses the top paired and unpaired methods by substantial margins. Fig.~\ref{fig3:dehaze} depicts visual comparisons with PM-Net~\cite{liu2022phase}, RIDCP~\cite{wu2023ridcp}, NSDNet~\cite{fan2023non}, MAP-Net~\cite{xu2023video}, and DVD~\cite{fan2024driving}. While these methods generally yield visually appealing dehazing results, DCL restores clearer predictions with more accurate content and outlines. Additionally, it is capable of estimating valid depth, a feature not offered by other dehazing models.

\noindent\textbf{Monocular depth estimation.} 
To further access the performance of our DCL in depth estimation, we compare it against well-known self-supervised depth estimation approaches, including MonoDepth2~\cite{godard2019digging}, MonViT~\cite{zhao2022monovit}, RobustDepth~\cite{saunders2023self}, Mono-ViFI~\cite{liu2024mono}, and Lite-Mono~\cite{zhang2023lite}. Due to the scarcity of hazy data with depth annotations, we train these methods on the GoProHazy benchmark and evaluate them on the DENSE-Fog dataset. It is worth noting that, during the testing phase, our DCL only uses the depth estimation subnetwork. Table.~\ref{tab2:depth} reports the quantitative results. Overall, in both light and dense fog scenarios, our DCL outperforms the others across nearly all five evaluation metrics. In dense fog scenes, accuracy and error metrics show some inconsistency, as the blurred depth estimates tend to be closer to the mean of the ground truth, resulting in smoother predictions. Figure~\ref{fig4:depth} illustrates the visual results for MonoViT, Mono-ViFI, and DCL. As shown, the depth predictions from MonoViT and Mono-ViFI are blurry and unreliable, while our DCL generates more accurate depth estimates and also provides cleaner dehazed images.

\begin{table}[!t]
	\setlength\tabcolsep{1.2pt}
	\renewcommand\arraystretch{1.2}
	\centering
	\scalebox{0.80}{
		\begin{tabular}{l||c|c|c||c|c|c}
			\hline
			Method
			& \multicolumn{1}{c|}{BCC} 
			& \multicolumn{1}{c|}{$D_{MFIR}$} 
			& \multicolumn{1}{c||}{$D_{MDR}$} 
			& \multicolumn{1}{c|}{Abs Real$\downarrow$} 
			& \multicolumn{1}{c|}{RMSE log$\downarrow$} 
			& \multicolumn{1}{c}{$\delta_1$$\uparrow$} \\
			\hline
			DCL w/o BCC 		 &              & $\checkmark$ & $\checkmark$      &0.636		&0.569       &0.439         \\
			DCL w/o $D_\text{MFIR}$ & $\checkmark$ &              & $\checkmark$      &0.320		&0.366       &0.621        	\\
			DCL w/o $D_\text{MDR}$  & $\checkmark$ & $\checkmark$ &                   &0.340		&0.392       &0.562         \\
			\hdashline
			\textbf{DCL (Ours)}& $\checkmark$ & $\checkmark$ & $\checkmark$      &{\bf 0.311}	&{\bf 0.364} &{\bf 0.623}   \\
			\hline
	\end{tabular}}%
    \vskip -0.05in
	\caption{Ablation study on DENSE-Fog (light).}
	\label{tab3:ablation_module}%
    \vskip -0.05in
\end{table}

\noindent\textbf{Model Efficiency.}
We compared the parameter count, FLOPs, and inference time of the SOTA methods for image/video dehazing and self-supervised depth estimation tasks on an NVIDIA RTX 4090 GPU. The running time was measured with an input size of 640$\times$192. As shown in Table~\ref{tab1:dehaze} and Table.~\ref{tab2:depth}, Our method achieved the shortest inference times of 0.075s and 0.009s for image/video dehazing and self-supervised depth estimation tasks, respectively, demonstrating that DCL offers fast inference performance.

\begin{figure}[!t]
	\centering
	\includegraphics[width=0.95\linewidth]{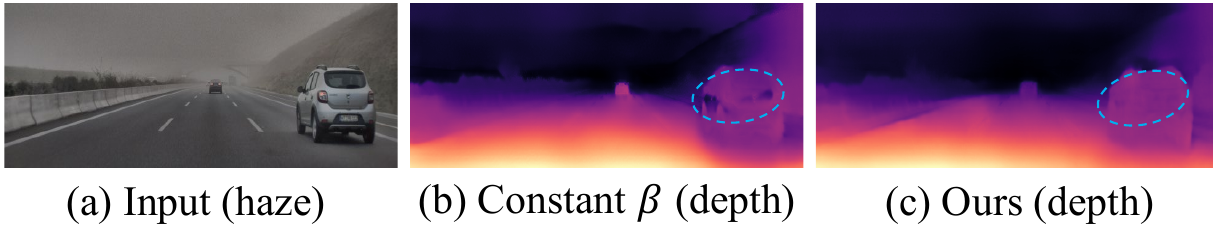}
    \vskip -0.05in
	\caption{Visual comparison of depth estimation across different $\beta$ types on DENSE-Fog (light).}
	\label{fig6:diffent_beta}
    \vskip -0.05in
\end{figure}

\subsection{Ablation Study}\label{ablation study}

\noindent\textbf{Effect of BCC, $D_{\text{MFIR}}$ and $D_{\text{MDR}}$.} To evaluate the effectiveness of our proposed BCC, $D_{\text{MFIR}}$ and $D_{\text{MDR}}$, we conducted experiments by excluding each component and training our model. The results in Table \ref{tab3:ablation_module} and Fig. \ref{fig5:ablation_module} demonstrate a significant improvement in video dehazing when the BCC module is integrated. This improvement is attributed to the use of dehazed images for enforcing a brightness consistency constraint, which leads to more accurate depth estimation ($d$) and more efficient haze removal as per Eq. \eqref{eq.asm}. Furthermore, both $D_{\text{MFIR}}$ and $D_{\text{MDR}}$ contribute to improvements in both dehazing and depth estimation.

\begin{table}[!t]
	\centering
	\setlength\tabcolsep{7.5pt}
	\renewcommand\arraystretch{1.2}
	\scalebox{0.80}{
		\begin{tabular}{l||c|c|c||c|c}
			\hline
			Method
			& \multicolumn{1}{c|}{$\mathcal{L}_{pe}$} 
			& \multicolumn{1}{c|}{$\mathcal{L}_{s}$} 
			& \multicolumn{1}{c||}{$\mathcal{L}_{rec}$} 
			& \multicolumn{1}{c|}{FADE$\downarrow$} 
			& \multicolumn{1}{c}{NIQE$\downarrow$} \\
			\hline
			DCL w/o $\mathcal{L}_{pe}$ 		&              & $\checkmark$ & $\checkmark$      & 0.6959 & 3.4785   \\
			DCL w/o $\mathcal{L}_{s}$       & $\checkmark$ &              & $\checkmark$      & 0.8163 & 3.5973   \\
			DCL w/o $\mathcal{L}_{rec}$     & $\checkmark$ & $\checkmark$ &                   & 0.7581 & 3.7030   \\
			\hdashline
			\textbf{DCL (Ours)}  		    & $\checkmark$ & $\checkmark$ & $\checkmark$      &{\bf 0.6914} &{\bf 3.4412}   \\
			\hline
	\end{tabular}}%
    \vskip -0.05in
	\caption{Ablation studies on different losses on GoProHazy.}
	\label{tab4:ablation_loss}%
    \vskip -0.05in
\end{table}%

\begin{table}[!t]
	\centering
	\setlength\tabcolsep{2.5pt}
	\renewcommand\arraystretch{1.3}
	\scalebox{0.80}{
		\begin{tabular}{c||c||c|c|c}
			\hline
			Shape of $\beta$ & Type & Abs Rel$\downarrow$ & RMSE log$\downarrow$ & $\delta_1$$\uparrow$    \\
			\hline
			(1, 1, 1)       & Constant       & 0.325     & 0.371     & 0.621       \\
			\hline
			\textbf{(1, 192, 640) (Ours)}   & Non-uniform    & {\bf0.311}     & {\bf0.364}     & {\bf0.623}   \\
			\hline
	\end{tabular}}%
    \vskip -0.05in
	\caption{Quantitative comparison of depth estimation across different $\beta$ types on DENSE-Fog (light).}
	\label{tab5:qualitative_beta}
    \vskip -0.05in
\end{table}

\noindent\textbf{Effect of the losses $\mathcal{L}_{pe}$, $\mathcal{L}_{s}$ and $\mathcal{L}_{rec}$.} We conducted a series of experiments to evaluate the effectiveness of the losses $\mathcal{L}_{pe}$, $\mathcal{L}_{s}$ and $\mathcal{L}_{rec}$ on GoProHazy. 
The FADE and NIQE results are reported in Table \ref{tab4:ablation_loss}. The findings clearly demonstrate that $\mathcal{L}_{rec}$ plays a pivotal role, as the ASM model, described in Eq. \eqref{eq.asm}, represents a key physical mechanism in video dehazing and ensures the independence of the dehazed results from the misaligned clear reference frame. Moreover, the smoothness loss $\mathcal{L}_{s}$ significantly contributes to both video dehazing and depth estimation. 

\noindent\textbf{Discussion on $\beta$ type.} In our methodology, we assume that $\beta$ is a non-uniform variable, as haze in real-world scenes is typically non-uniform, such as patchy haze. Consequently, scattering coefficients vary across different regions. While most existing works use a constant scattering coefficient, we perform comparative experiments with both constant and non-uniform $\beta$ values, as presented in Table~\ref{tab5:qualitative_beta}. The results demonstrate that using a non-uniform $\beta$ significantly improves depth estimation accuracy, which is further validated by visual comparisons in Fig.~\ref{fig6:diffent_beta}.

\section{Conclusion}\label{conclusion}
In this paper, we developed a new Depth-centric Learning framework (DCL) by proposing a unified ASM-BCC model that integrates the atmospheric scattering model with the brightness consistency constraint via a shared depth estimation network. This network leverages adjacent dehazed frames to enhance depth estimation using BCC, while refined depth cues improve haze removal through ASM. Furthermore, we utilize a misaligned clear video and its estimated depth to regularize both the dehazing and depth estimation networks with two discriminator networks: $D_\text{MFIR}$for enhancing high-frequency details and $D_\text{MDR}$ for mitigating black hole artifacts. Our DCL framework outperforms existing methods, achieving significant improvements in both video dehazing and depth estimation in real-world hazy scenarios.

\section{Acknowledgments}
This work was supported by the National Science Fund of China under Grant Nos. U24A20330, 62361166670, and 62072242.

{\bibliography{aaai25}}

\clearpage
\appendix
\setcounter{page}{1}

\setcounter{table}{0}
\setcounter{figure}{0}
\setcounter{equation}{0}
\setcounter{section}{0}
\renewcommand\thesection{\Alph{section}}
\renewcommand{\thetable}{S\arabic{table}}
\renewcommand{\thefigure}{S\arabic{figure}}
\renewcommand{\theequation}{S\arabic{equation}}

\twocolumn[{
	\section*{\huge{Supplemental Material}}
	\vspace{10mm}
}]

In this supplementary material, we provide an experiment on the REVIDE~\cite{zhang2021learning} dataset in Sec. \textbf{A} and more implementation details~\ref{more implementation details} in Sec. \textbf{B}. Next, we present depth evaluation metrics in Sec. \textbf{C} and include additional ablation studies and discussions in Sec. \textbf{D}. In Sec. \textbf{E}, we showcase more visual results, including video dehazing and depth estimation results.  \\

\section{A. Experiment on REVIDE dataset.}

\begin{figure}[!ht]
	\centering
	\includegraphics[width=0.90\linewidth]{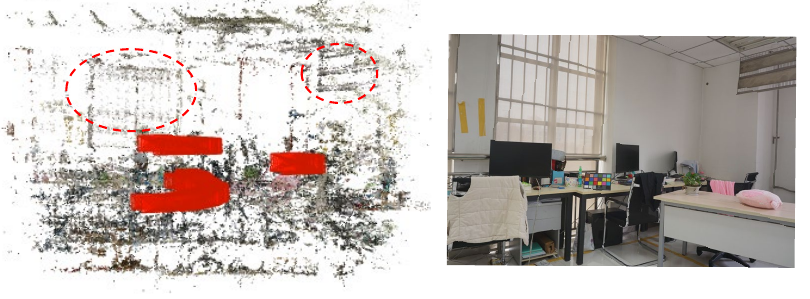}
	\caption{Use COLMAP to calibrate the camera intrinsics of the REVIDE dataset.} 
	\label{figS1:camera calibrate}
\end{figure}

\noindent \textbf{Camera calibration.} Since the REVIDE dataset does not provide camera intrinsics, we selected a high-quality continuous indoor video and used COLMAP for 3D reconstruction to obtain the camera intrinsics. The resulting camera trajectory aligns with the movement path of the robotic arm.

 \begin{table}[!ht]
	\centering
	\setlength\tabcolsep{1.5pt}
	\renewcommand\arraystretch{1.2}
	\scalebox{0.84}{
		\begin{tabular}{l|c|c|c|c|c}
			\hline
			\multicolumn{1}{c|}{\multirow{2}[2]{*}{\makecell{Data\\Settings}}} 
			& \multirow{2}[2]{*}{Methods} & \multicolumn{2}{c|}{REVIDE} 
			& \multicolumn{1}{c|}{\multirow{2}[2]{*}{\makecell{Inf. time\\(s)}}} 
			& \multirow{2}[1]{*}{Ref.} \\
			\cline{3-4} &   & \multicolumn{1}{c|}{PSNR $\uparrow$} & \multicolumn{1}{c|}{SSIM $\uparrow$} & & \\
			\hline
			\multicolumn{1}{c|}{\multirow{4}[2]{*}{\makecell{Unpaired}}} & DCP  & 11.03  & 0.7285 & 1.39 & CVPR'09 \\
			& RefineNet   & 23.24  & 0.8860 & 0.105 & TIP'21 \\
			& CDD-GAN    & 21.12  & 0.8592 & 0.082 & ECCV'22 \\
			& D$^{4}$   & 19.04  & 0.8711  & 0.078 & CVPR'22 \\
			\hdashline
			\multicolumn{1}{c|}{\multirow{2}[9]{*}{\makecell{Paired}}} & PSD  & 15.12  & 0.7795 & 0.084 & CVPR'21 \\
			& RIDCP  & 22.70  & 0.8640 & 0.720  & CVPR'23 \\
			& PM-Net  & 23.83  & 0.8950  & 0.277 & ACMM'22 \\
			& MAP-Net  & 24.16  & 0.9043  & 0.668 & CVPR'23 \\
			\hdashline
			\multirow{2}[5]{*}{Non-aligned} & NSDNet  & 23.52   & 0.8892  & {\bf 0.075}  & arXiv'23 \\
			&DVD  & 24.34  & 0.8921 & 0.488 & CVPR'24 \\
			&\textbf{DCL (ours)}  & {\bf 24.52}  &{\bf 0.9067} & {\bf 0.075} & - \\
			\hline
	\end{tabular}}%
	\caption{Comparison of the proposed method and methods with aligned ground truth on REVIDE dataset.}
	\label{tabS1: REVIDE}%
\end{table}%

\begin{figure}[!ht]
	\centering
	\includegraphics[width=0.90\linewidth]{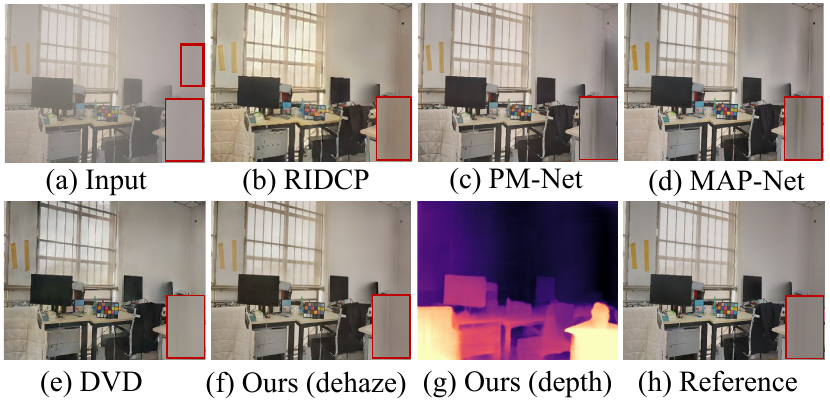}
	\caption{Visual comparison on REVIDE dataset.} 
	\label{figS2:REVIDE}
     \vskip -0.1in
\end{figure}

\noindent\textbf{Evaluation on REVIDE.} To further assess the effectiveness of our proposed method, we evaluate all state-of-the-art (SOTA) dehazing methods on the real smoke dataset with ground truth (REVIDE) using PSNR and SSIM metrics. As shown in Table~\ref{tabS1: REVIDE}, our proposed method achieves the highest values. In this work, we primarily focus on video dehazing and depth estimation in real driving scenarios. However, we also obtain excellent experimental results on the real smoke dataset, indicating that our method is effective for smoke removal. Additionally, we provide visual comparisons in Fig.~\ref{figS2:REVIDE}, where the dehazing results of all competing methods exhibit artifacts and suboptimal detail restoration. In contrast, the proposed method generates much clearer results that are visually closer to the ground truth.

\section{B. More Implementation Details}\label{more implementation details}

For the depth and pose estimation networks, we mainly followed the depth and pose architecture of Monodepth2~\cite{godard2019digging}. Our dehaze network is an encoder-decoder architecture without skip connections. This network consists of three convolutions, several residual blocks, two fractionally strided convolutions with stride 1/2, and one convolution that maps features to $\mathbb{R}^{3\times 256\times 256}$, and we use 9 residual blocks. For the D$_\text{MFIR}$ and D$_\text{MDR}$ discriminator networks, we use 70$\times$70 PatchGAN~\cite{zhu2017unpaired} network. Additionally, our scatter coefficient estimation network is similar to the depth estimation network, and shares the encoder network weights.

\section{C. Depth Evaluation Metrics}\label{depth evaluation metrics}
Five standard metrics are used for evaluation, including Abs Rel, RMSE log, $\delta_1$, $\delta_2$ and $\delta_3$, which are presented by
\begin{equation}
	\begin{split}
		\text{Abs Rel}=\frac{1}{|D|}\sum\nolimits_{d^*\in D}{|d^*-d|/d^*},\\
		\text{RMSE log}=\sqrt{\frac{1}{|D|}\sum\nolimits_{d^*\in D}{\|logd^*-logd\|^2}}, \\
		\delta_i=\frac{1}{|D|}|\{d^*\in D \max (\frac{d^*}{d},\frac{d}{d^*})<1.25^i \}|,
	\end{split}
	\label{equ.metrics}
\end{equation}

where $d$ and $d^*$ denote predicted and ground truth depth maps, respectively. $D$ represents a set of valid ground truth depth values in one image, and $|\cdot|$ returns the number of elements in the input set.

\begin{figure}[!ht]
	\centering
	\includegraphics[width=0.90\linewidth]{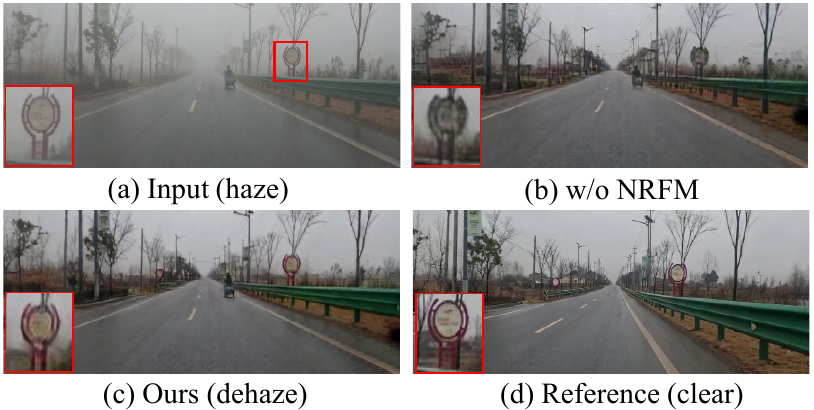}
	\caption{Ablation visualization for NRFM. The NRFM significantly enhances the dehazing results.} 
	\label{figS3:ablation_NRFM}
\end{figure}

\begin{table}[!ht]
	\centering
	\setlength\tabcolsep{6.5pt}
	\renewcommand\arraystretch{1.3}
	\scalebox{0.8}{
		\begin{tabular}{l||c|c}
			\hline
			Model & FADE$\downarrow$  & NIQE$\downarrow$ \\
			\hline
			DCL wo / NRFM           & 0.7216     	    & 3.4766 \\
			\hline
			\textbf{DCL (Ours)}     & {\bf0.6914}     & {\bf3.4412} \\
			\hline
	\end{tabular}}%
	\caption{Ablation studies for the NRFM on GoProHazy}
	\label{tabS2:ablation_NRFM}%
     \vskip -0.1in
\end{table}

\section{D. More Ablation and Discussions}\label{more ablation studies}

\textbf{Effect of different input frames.} Tab.~\ref{tabS3:different_inputframes} shows that using a 3-frame input yields the best results, with only a small difference in quantitative results between 2-frame and 3-frame inputs. By using a 3-frame input, occlusion is alleviated, resulting in sharper depth results. Here, to obtain the best experimental results, we use 3 frames as the input for our model.

\noindent\textbf{Effect of NRFM.} To evaluate the effect of NRFM, we conducted experiments without the NRFM module, training our model in an unpaired setting with randomly matched clear reference frames. The results in Tab.~\ref{tabS2:ablation_NRFM} and Fig.~\ref{figS3:ablation_NRFM} show a significant enhancement in video dehazing with the NRFM module. This improvement is attributed to a more robust supervisory signal derived from misaligned clear reference frames, which is distinct from the unpaired setting. 

\begin{table}[!ht]
	\setlength\tabcolsep{2.5pt}
	\renewcommand\arraystretch{1.5}
	\centering
	\scalebox{0.8}{
	\begin{tabular}{l||c|c|c|c|c|c}
		\hline
		Input frames & $\#$ Number& Abs Rel$\downarrow$ & RMSE log$\downarrow$ & $\delta_1$$\uparrow$  & $\delta_2$$\uparrow$ & $\delta_3$$\uparrow$ \\
		\hline
		$\left[-1, 0\right]$      &2  & 0.316        &  {\bf0.364}   & {\bf0.623}    & {\bf0.839}   &  {\bf0.921}  \\
		\hline
		$\left[-1, 0, 1\right]$   &3  & {\bf0.311}   & {\bf0.364}    &  {\bf0.623}   & {\bf0.839}   & {\bf0.921} \\
		\hline
	\end{tabular}}%
	\caption{Ablation studies for the number of input frames on DENSE-Fog (light) dataset.}
	\label{tabS3:different_inputframes}%

\end{table}

\begin{figure}[!t]
	\centering
	\includegraphics[width=0.90\linewidth]{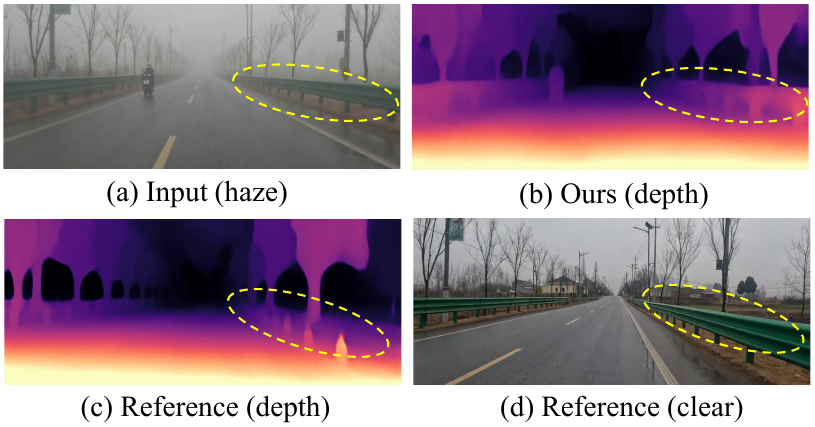}
	\caption{Visualizing the gains brought by ASM constraints to depth estimation.} 
	\label{figS4:our_depth_VS_ref_depth}
\end{figure}

\noindent\textbf{Discussion on predicted depth surpassing reference depth.} As shown in Fig.~\ref{figS4:our_depth_VS_ref_depth}, we visually compared the predicted depth with the reference depth. The experimental results indicate that the predicted depth is superior to the reference depth, primarily due to the significant constraint imposed by the atmospheric scattering model through the reconstruction loss ($\mathcal{L}_{rec}$) on depth estimation. Additionally, this indirectly verifies that our method outperforms the two-stage depth estimation approach (i.e., dehazing first, then depth estimation) for hazy scenes, as the reference depth comes from Monodepth2~\cite{godard2019digging}, which was trained on clear, misaligned reference videos.

\section{E. More Visual Results}\label{more visual results}

\begin{figure*}[!ht]
	\centering
	\includegraphics[width=1.0\linewidth]{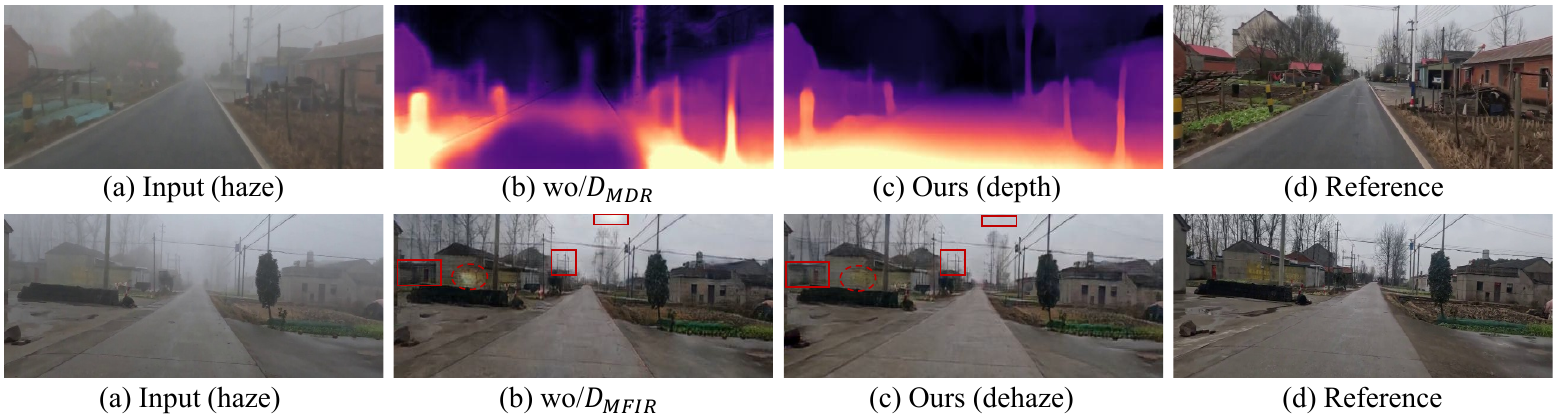}
	\caption{Ablation visualizations of $D_\text{MDR}$ and $D_\text{MFIR}$ are shown respectively on the depth and dehaze results from the GoProHazy dataset.} 
	\label{figS9:ablation}
\end{figure*}

\textbf{More visualizations of ablation studies.} As shown in Fig.~\ref{figS9:ablation}, we visualize the ablation of the proposed $D_\text{MDR}$ in weak texture road surface scenarios to highlight its advantages. We also showcase the proposed $D_\text{MFIR}$ on dehaze results to emphasize its impact on texture details.

\begin{figure*}[!ht]
	\centering
	\subfigure{
		\begin{minipage}[b]{1.0\textwidth}	
			\includegraphics[width=1\textwidth]{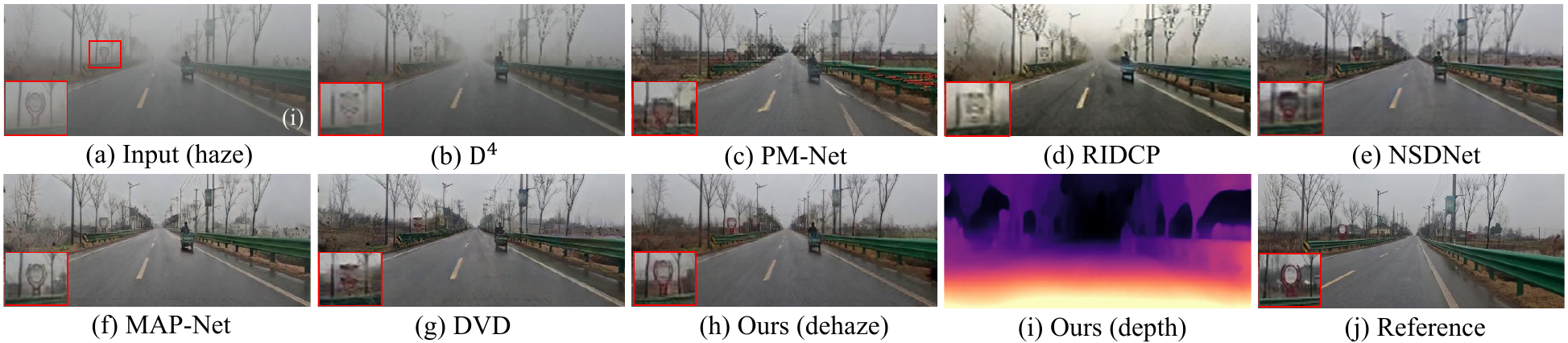}	
	\end{minipage}}
	\subfigure{
		\begin{minipage}[b]{1.0\textwidth}
			\includegraphics[width=1\textwidth]{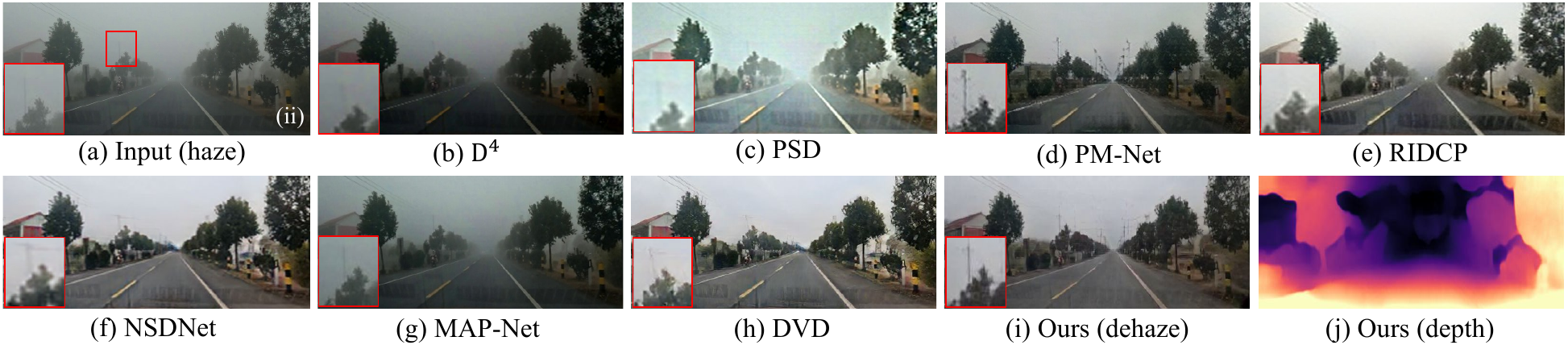}
	\end{minipage}}
	\subfigure{
		\begin{minipage}[b]{1.0\textwidth}
			\includegraphics[width=1\textwidth]{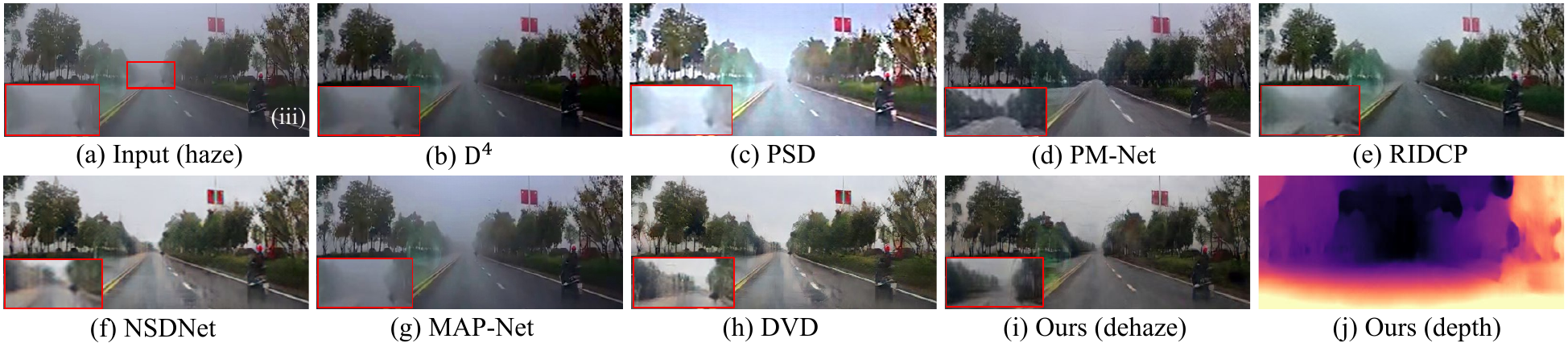}
	\end{minipage}}
	\caption{Comparing video dehazing results on GoProHazy (i), DrivingHazy (ii), and InternetHazy (iii), respectively, our method effectively removes haze and estimates depth.}
	\label{figS5:dehaze}
\end{figure*}

\noindent\textbf{More visualizations of video dehazing.} In Fig.~\ref{figS5:dehaze}, we show additional visual comparison results with state-of-the-art image/video dehazing methods on the GoProHazy dataset (i), DrivingHazy dataset (ii) and InternetHazy dataset (iii), respectively. The visual comparison results demonstrate that our proposed DCL method performs better in video dehazing, particularly in distant haze removal and the restoration of close-range texture details.

\begin{figure*}[!ht]
	\centering
	\includegraphics[width=1.0\linewidth]{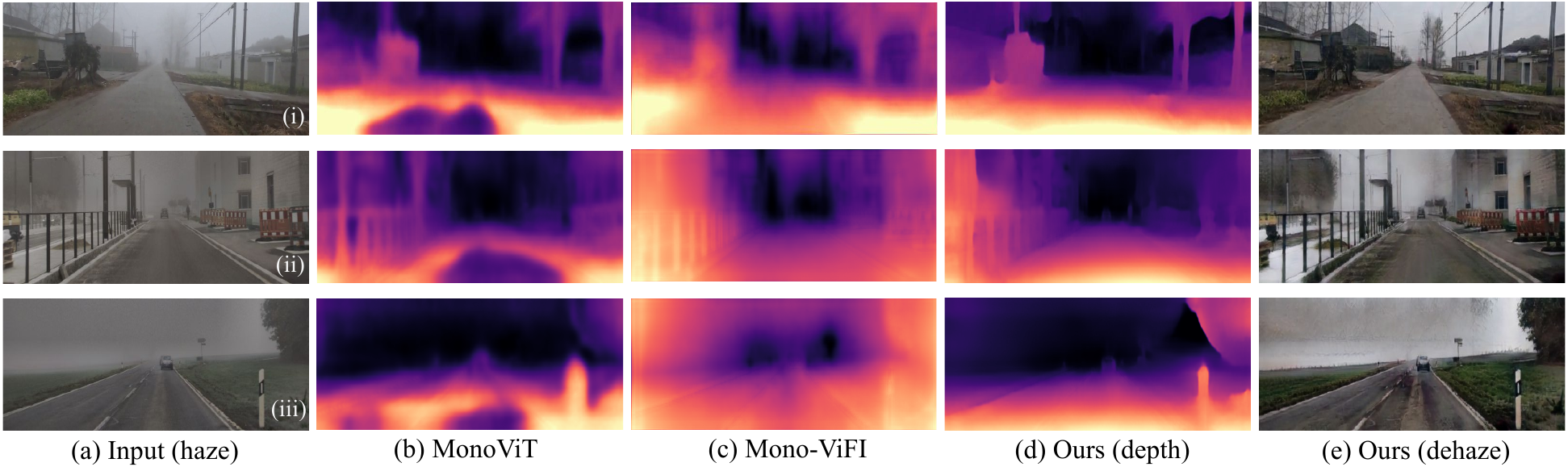}
	\caption{Visual comparison on GoProHazy (i) and DENSE-Fog (ii-dense, iii-light). Our method can estimate more accurate depth in real hazy scenes.} 
	\label{figS6:depth_1}
\end{figure*}

\begin{figure*}[!ht]
	\centering
	\includegraphics[width=1.0\linewidth]{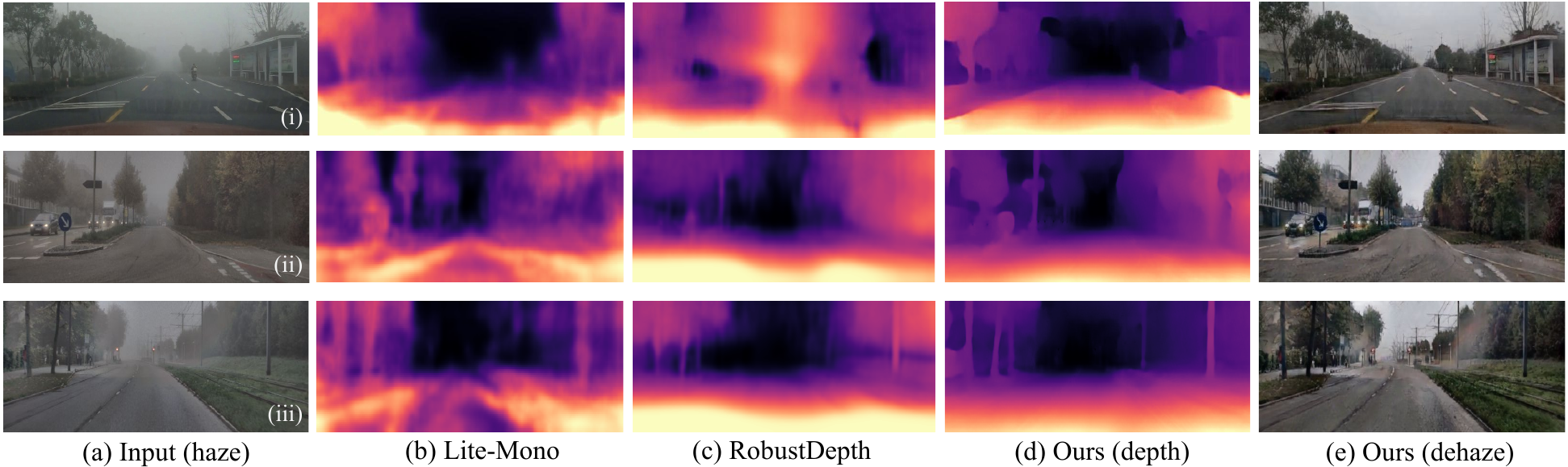}
	\caption{Visual comparison on DrivingHazy (i) and DENSE-Fog (ii-dense, iii-light). Our method can estimate more accurate depth in real hazy scenes.} 
	\label{figS7:depth_2}
\end{figure*}

\noindent\textbf{More visualizations of depth estimation.} We separately present additional visual comparison results with state-of-the-art depth estimation methods on the GoProHazy(i), DENSE-Fog(dense-(ii) and light-(iii)), as shown in Fig.~\ref{figS6:depth_1} and Fig.~\ref{figS7:depth_2}. Mono-ViFI~\cite{liu2024mono}, Lite-Mono~\cite{zhang2023lite}, and RobustDepth~\cite{saunders2023self} produce blurry and inaccurate depth maps. Additionally, MonoViT appears to have good clarity, but it creates black holes in areas of weak texture on the road. In contrast, our DCL still provides a plausible prediction.

\begin{figure*}[!ht]
	\centering
	\includegraphics[width=1.0\linewidth]{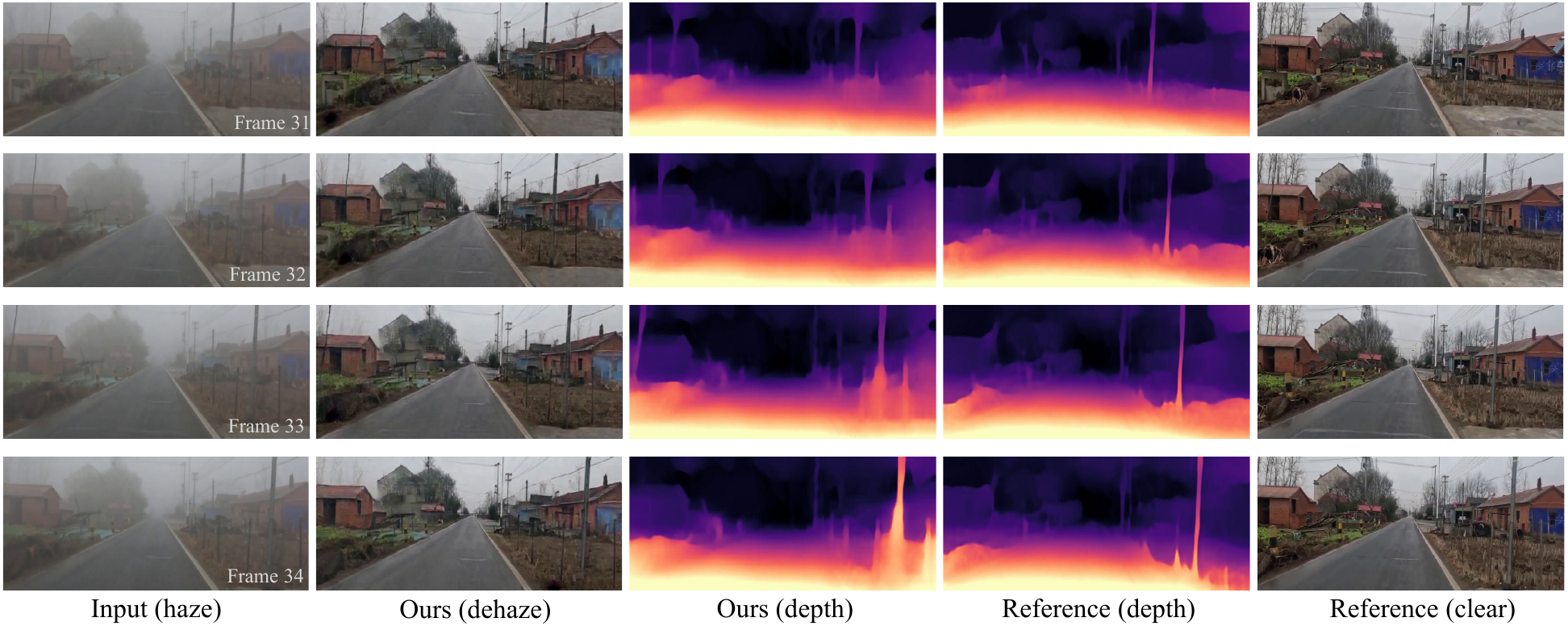}
	\caption{Our method visualizes consecutive frame dehazing and depth estimation results on GoProHazy.} 
	\label{figS8:our_dehaze_depth}
\end{figure*}

\noindent\textbf{More visualization of DCL on video dehazing and depth estimation.} To validate the stability of our proposed DCL in video dehazing and depth estimation, we showcase the consecutive frame dehazing results and depth estimation on the GoProHazy dataset in Fig.~\ref{figS8:our_dehaze_depth}. The visual results show that our method maintains good brightness consistency between consecutive frames, especially in the sky region. It effectively removes distant haze and restores texture details without introducing artifacts in the sky. Additionally, our method avoids incorrect depth estimations in areas with weak textures on the road surface, such as black holes.

\noindent\textbf{Video demo.} To demonstrate the stability of the proposed method, we separately compared it with the latest state-of-the-art video dehazing (e.g., MAP-Net~\cite{xu2023video}, DVD~\cite{fan2024driving}) and monocular depth estimation methods (e.g., Lite-Mono) on GoProHazy. We have included the \textbf{video-demo.mp4} file in the supplementary materials.

\noindent\textbf{Limitations.} In real dense hazy scenarios, our method struggles to recover details of small objects, such as tree branches and wires, which can often result in artifacts in the dehazed images. This is mainly due to the difficulty of effectively extracting such subtle feature information for the network. Moreover, obtaining a large amount of high-quality misaligned data in dynamic scenes with people and vehicles is challenging, which limits the model's generalization ability.

\vskip 5in

\end{document}